%% file: main.tex
\documentclass[runningheads]{llncs}

\usepackage[T1]{fontenc}
\usepackage{graphicx}
\usepackage{multirow}
\usepackage{booktabs}
\usepackage{amsfonts}
\usepackage{amssymb}
\usepackage{amsmath}
\usepackage{dsfont}
\usepackage{ctable}
\usepackage{color, colortbl, xcolor, soul}
\usepackage[colorlinks=true,linkcolor=blue, citecolor=blue, urlcolor=blue]{hyperref}
\usepackage{pifont}
\usepackage{etoolbox}

\let\oldnl\nl
\newcommand\scalemath[2]{\scalebox{#1}{\mbox{\ensuremath{\displaystyle #2}}}}
\newcommand{\nonl}{\renewcommand{\nl}{\let\nl\oldnl}}
\newcommand{\RomanNumeralCaps}[1]{\MakeUppercase{\romannumeral #1}}
\newcommand{\cmark}{\ding{51}}

\definecolor{Gray1}{gray}{0.95}
\definecolor{Gray2}{gray}{0.88}
\definecolor{Gray3}{gray}{0.81}

\definecolor{babyblueeyes}{rgb}{0.63, 0.79, 0.95}
\sethlcolor{babyblueeyes}

\begin{document}
\title{MAP: Domain Generalization via \underline{M}eta-Learning on \underline{A}natomy-Consistent \underline{P}seudo-Modalities}
\titlerunning{MAP}

\author{Dewei Hu\inst{1}\and
Hao Li\inst{1} \and
Han Liu\inst{2} \and
Xing Yao\inst{2} \and
Jiacheng Wang\inst{2} \and
Ipek Oguz\inst{1}\inst{2}}

\authorrunning{D. Hu et al.}

\institute{Department of Electrical and Computer Engineering, Vanderbilt University\and 
Department of Computer Science, Vanderbilt University\\
\email{dewei.hu@vanderbilt.edu}}

\maketitle          
\begin{abstract}
Deep models suffer from limited generalization capability to unseen domains, which has severely hindered their clinical applicability. Specifically for the retinal vessel segmentation task, although the model is supposed to learn the anatomy of the target, it can be distracted by confounding factors like intensity and contrast. We propose Meta learning on Anatomy-consistent Pseudo-modalities (MAP), a method that improves model generalizability by learning structural features. We first leverage a feature extraction network to generate three distinct pseudo-modalities that share the vessel structure of the original image. Next, we use the episodic learning paradigm by selecting one of the pseudo-modalities as the meta-train dataset, and perform meta-testing on a continuous augmented image space generated through Dirichlet mixup of the remaining pseudo-modalities. Further, we introduce two loss functions that facilitate the model's focus on shape information by clustering the latent vectors obtained from images featuring identical vasculature. We evaluate our model on seven public datasets of various retinal imaging modalities and we conclude that MAP has substantially better generalizability. Our code is publically available at \url{https://github.com/DeweiHu/MAP}.

\end{abstract}
\section{Introduction}
\begin{figure}[t]
    \newcommand{\rsize}{0.175}
    \setlength{\tabcolsep}{2.5pt}
    \centering
    \begin{tabular}{ccccc}
        \includegraphics[width=\rsize\linewidth]{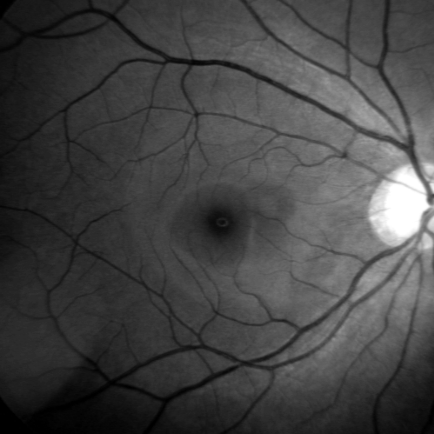} &  \includegraphics[width=\rsize\linewidth]{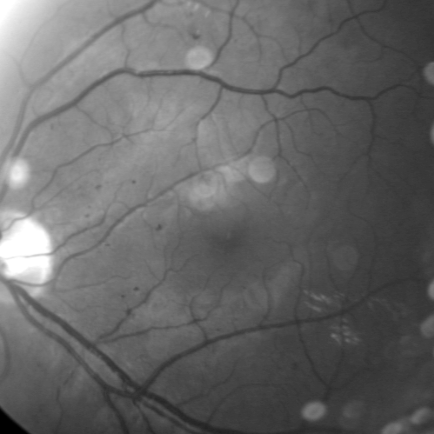} &
        \includegraphics[width=\rsize\linewidth]{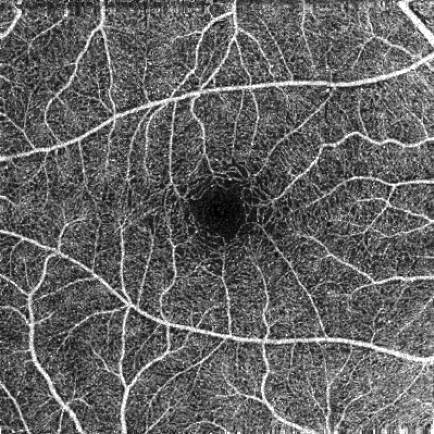} &
        \includegraphics[width=\rsize\linewidth]{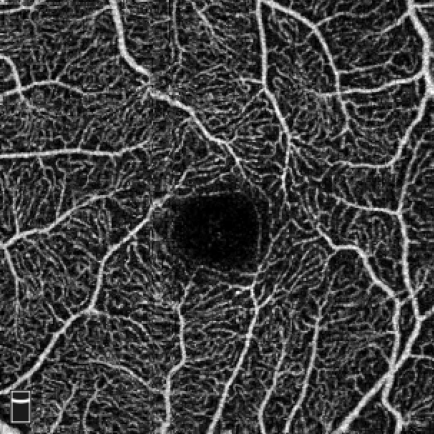} &
        \includegraphics[width=\rsize\linewidth]{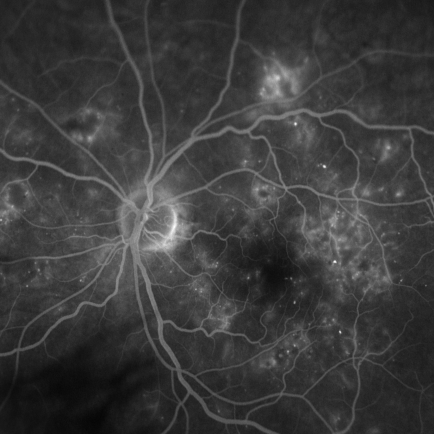} \\
        \scriptsize(a)healthy fundus & \scriptsize(b)diseased fundus &\scriptsize (c)OCT-A site-1 & \scriptsize(d)OCT-A site-2 & \scriptsize(e) FA
    \end{tabular}
    \caption{Domain shift examples. Type \RomanNumeralCaps{1}: pathological phenotypes (a vs.\ b). Type \RomanNumeralCaps{2}: cross-site shifts (c vs.\ d). Type \RomanNumeralCaps{3}: cross-modality shifts (a-b vs.\ c-d vs.\ e).}
    \label{fig:data_example}
\end{figure}

In the absence of a single standardized imaging paradigm, medical images obtained from different devices may exhibit considerable domain variation. Fig.\ \ref{fig:data_example} demonstrates three types of domain shift among images delineating the retinal vessels. The presence of such distribution mismatch can significantly degrade the performance of deep learning models on unseen datasets, thus impeding their widespread clinical deployment. To address the domain generalization (DG) problem \cite{zhou2022domain}, a straightforward idea is to focus on the domain-invariant patterns for the specific downstream task. For retinal vessel segmentation, the morphology of vessels can be deemed such a domain-invariant pattern. Hence, our hypothesis is that \textit{emphasizing the structural characteristics of the vasculature can enhance the model's DG performance}. Following a similar idea, Hu et al.\ \cite{hu2022domain} proposed to explicitly delineate the vessel shape by a Hessian-based vector field. However, the dependency on the image gradient makes this approach vulnerable to low-quality data with poor contrast and/or high noise. In contrast, we instead propose an implicit way of exploiting the morphological features by adopting the \underline{\textbf{m}}eta-learning paradigm on \underline{\textbf{a}}natomy-consistent \underline{\textbf{p}}seudo-modalities (MAP).

\begin{figure}[t]
    \centering
    \includegraphics[width=.95\linewidth]{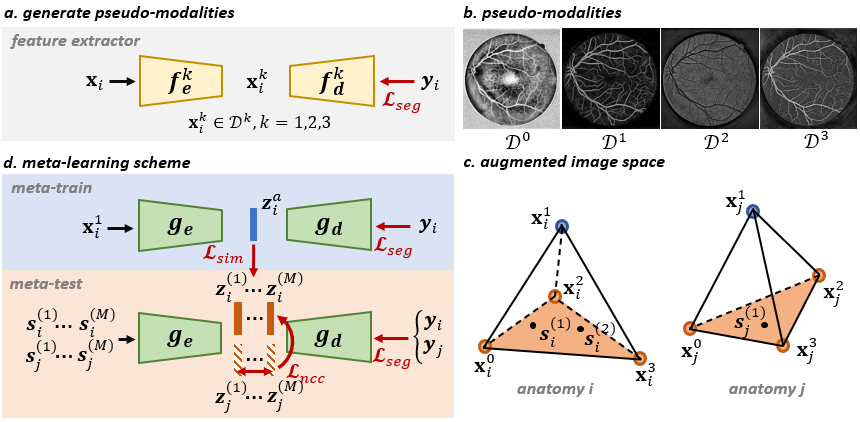}
    \caption{The key components of MAP, clockwise. \textbf{(a)} $f(\cdot)$ is the synthesis network. $\mathbf{x}_i$ is the $i^{th}$ color fundus input and $\mathbf{y}_i$ is its  ground truth vessel map. $k$ indexes three different models that generate diverse pseudo-modalities. \textbf{(b)} An example image in four pseudo-modalities: $\mathcal{D}^0$ is the histogram equalization of intensity-reversed green channel of input $\mathbf{x}$ and $\mathcal{D}^k$, $k=1,2,3$ are generated by $f_e^k$. \textbf{(c)} The four pseudo-modalities of an input $\mathbf{x}_i$ form the corners of a tetrahedron. The colored facet is a continuous image space created by Dirichlet mixup. $\mathbf{s}_{i}^{(m)}$ denotes the $m^{th}$ sample from the image space. Anatomy $i$ represents the underlying shape of vasculature in $\mathbf{x}_i$, which is consistent for all samples $\mathbf{s}_{i}^{(m)}$. \textbf{(d)} The meta-learning scheme. $g(\cdot)$ is the segmentation network, $M$ is the number of samples drawn, $\mathbf{z}$ is the latent feature vector.}
    \label{fig:pipeline}
\end{figure}

First, we leverage a structural feature extraction network (Fig.\ \ref{fig:pipeline}(a)) generate three pseudo-modalities, similar to \cite{hu2022domain}. The network is defined by setting the bottleneck of the U-Net \cite{ronneberger2015u} backbone to have the same width and height with the input image. Given its capability to extract interpretable visualization, this architecture is often implemented in representation disentanglement \cite{ouyang2021representation} and unsupervised segmentation \cite{hu2021life}. Supervised by the binary vessel map, the latent image preserves the vasculature structure while the style exhibits some randomness, as illustrated in Fig.\ \ref{fig:pipeline}(b). Therefore, we refer to these latent images as  anatomy-consistent pseudo-modalities.  

Meta-learning has recently emerged as a popular technique for addressing  the DG problem \cite{dou2019domain,khandelwal2020domain}. Following the idea of episodic training presented in MAML \cite{finn2017model}, researchers split their training data into two subsets, meta-train and meta-test, to mimic the scenario of encountering out-of-distribution (OOD) data during training. Liu et al.\ \cite{liu2021feddg} proposed to conduct meta-learning in a continuous frequency space created by mixing up \cite{zhang2017mixup,kim2020puzzle} the amplitude spectrum. They keep the phase spectrum unchanged to preserve the anatomy in the generated images. In contrast, given our pseudo-modalities with identical underlying vasculature, we are able to create a continuous image space via Dirichlet mixup \cite{shu2021open} without affecting the vasculature. We regard images in each pseudo-modality as a corner of a tetrahedron, as depicted in Fig.\ \ref{fig:pipeline}(c). The red facet of the tetrahedron is a continuous space created by the convex combination of images from the three pseudo-modalities. We use images in one pseudo-modality (blue node) for meta-train and the mixup space (red facet) for meta-test. An important property of the mixup space is that all the samples share the same vessel structure while the image style may differ drastically. Hence, employing proper constraints on the relationship between features can implicitly encourage the model to learn the shape of vessels. Inspired by \cite{dou2019domain}, we leverage a similarity loss to express the feature consistency between the meta-train and meta-test stages. Additionally, we propose a normalized cross-correlation (NCC) loss to differentiate latent features extracted from images with different anatomy. In the context of contrastive learning, these loss functions cluster positive pairs and separate negative pairs. 

In our study, we use seven public datasets including color fundus, OCT angiography (OCT-A) and fluorescein angiography (FA) images. We train MAP on fundus data and test on all modalities. We show that MAP exhibits outstanding generalization ability in most conditions. Our main contributions are: 

\begin{itemize}
    \item[\ding{118}] We generate a continuous space of anatomy-consistent pseudo-modalities  with Dirichlet mixup.
    \item[\ding{118}] We present an episodic learning scheme employed on synthesized images.
    \item[\ding{118}] We propose a normalized cross-correlation loss function to cluster the feature vectors with regard to the vessel structure.
    \item[\ding{118}] We conduct extensive experiments on seven public datasets in various modalities which show the superior DG performance of MAP.
\end{itemize}      

\section{Methods}
\subsection{Problem Definition}
Given a source domain $\mathcal{S}=\{(\mathbf{x}_i,\mathbf{y}_i)|i\in\{1,\cdots,N\}\}$ that includes $N$ pairs of raw images $\mathbf{x}_i$ and ground truth labels $\mathbf{y}_i$, our goal is to train a segmentation network $g(\cdot)$ that can robustly work on the target domain $\mathcal{T}=\{T^{p}|p\in\{1,\cdots,P\}\}$ with $P$ unseen datasets. In practice, we include only fundus images in $\mathcal{S}$ since there are many public annotated fundus datasets. For $\mathcal{T}$, data from three different modalities (fundus, OCT-A and FA) are included. We test the model generalization on datasets with three distinct types of domain shift: (\RomanNumeralCaps{1}) data with pathological phenotypes, (\RomanNumeralCaps{2}) cross-site shifts, (\RomanNumeralCaps{3}) cross-modality shifts.

\subsection{Pseudo-modality Synthesis} \label{sec:synthesis}
The features in the latent space of a U-Net \cite{ronneberger2015u} backbone is usually a low-dimensional representation of the input images. In some applications (e.g., representation disentanglement), it is desirable for the latent features to show visually intuitive structural characteristics. In such scenarios, the bottleneck of the feature extraction network is set to have the same width and height as the input image. We adopt the approach presented in \cite{hu2022domain} to synthesize pseudo-modalities by exploiting this idea (Fig.\ \ref{fig:pipeline}(a)). Both the encoder $f_{e}$ and the decoder $f_{d}$ are residual U-Nets. The input $\mathbf{x}_i\in \mathbb{R}^{3\times H\times W }$ is a color image while $\mathbf{y}_i$ is the binary vessel map. The model is trained by optimizing a segmentation loss which is the sum of cross-entropy and the Dice loss \cite{ma2021loss}, i.e., $\mathcal{L}_{seg}=\mathcal{L}_{CE}+\mathcal{L}_{Dice}$. Without direct supervision, the latent image $\mathbf{x}_i^k$ can have a different appearance when the model is re-trained. Such randomness is purely introduced by the stochastic gradient descent (SGD) in the optimization process. $k=1,2,3$ indexes three different models and their corresponding synthesized image. For a fair comparison, we use the pre-trained models provided in \cite{hu2022domain} to generate the three pseudo-modalities ($\mathcal{D}^1$, $\mathcal{D}^2$, and $\mathcal{D}^3$) illustrated in Fig.\ \ref{fig:pipeline}(b). 

An essential property of the generated images is that despite significant intensity variations, they consistently maintain the shared anatomical structure of the vasculature. Therefore, the $\mathcal{D}^k$ are termed anatomy-consistent pseudo-modalities. To convert the input color fundus image $\mathbf{x}_i$ to grayscale, we conduct  histogram equalization (CLAHE) \cite{reza2004realization} on the intensity-reversed green channel and denote it as $\mathbf{x}_i^0$. The pseudo-modality of these pre-processed images is $\mathcal{D}^0$.

\subsection{Meta-learning on Anatomy Consistent Image Space} \label{sec:metalearning}
Developed from the few-shot learning paradigm, meta-learning seeks to enhance a model's generalizability to unseen data when presented with limited training sets. This is achieved by an episodic training paradigm that consists of two stages: meta-train and meta-test. The source domain $\mathcal{S}$ is split into two subsets $\mathcal{S}_{train}$ and $\mathcal{S}_{test}$ to mimic encountering OOD data during training. 

Mixup is a common strategy for data augmentation as it generates new samples via linear interpolation in either image \cite{kim2020puzzle} or feature space \cite{verma2019manifold}. Zhang et al.\ \cite{zhang2020does} showed Mixup improves model generalization and robustness. 
In \cite{liu2021feddg}, Liu et al.\ conduct meta-learning on generated images that are synthesized by mixing the amplitude spectrum in frequency domain. They preserve larger structures such as the optic disc by keeping the phase spectrum un-mixed. Given our anatomy-consistent pseudo-modalities, we are able to directly work on the images rather than the frequency domain. We select $\mathcal{D}^1$ as the meta-train data, and we mixup the remaining three pseudo-modalities ($\mathcal{D}^0$,$\mathcal{D}^2$, and $\mathcal{D}^3$) to form a continuous  space (red facet in Fig.\ \ref{fig:pipeline}(c)) from which we draw  meta-test samples.

\begin{figure}[t]
    \centering
    \begin{tabular}{cccccc}
        & $\boldsymbol\alpha=[5,5,5]$ & $\boldsymbol\alpha=[4,2,2]$ & $\boldsymbol\alpha=[1.5,5,5]$ & $\boldsymbol\alpha=[1.5,5,1.5]$&\\
        
        \rotatebox{90}{\hspace{0.6cm} $P(\boldsymbol\lambda)$} &
        \includegraphics[width=0.2\linewidth]{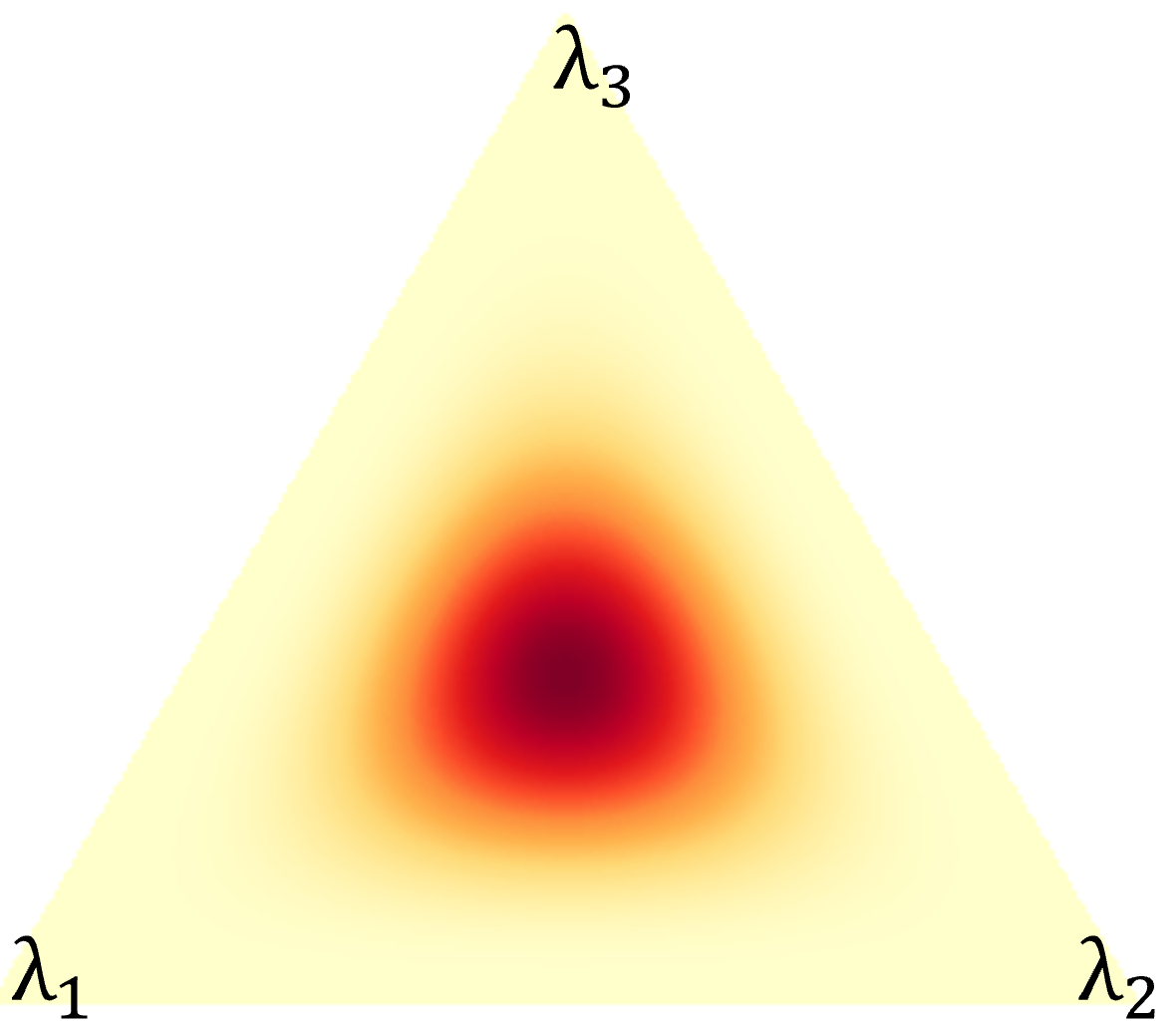}&  
        \includegraphics[width=0.2\linewidth]{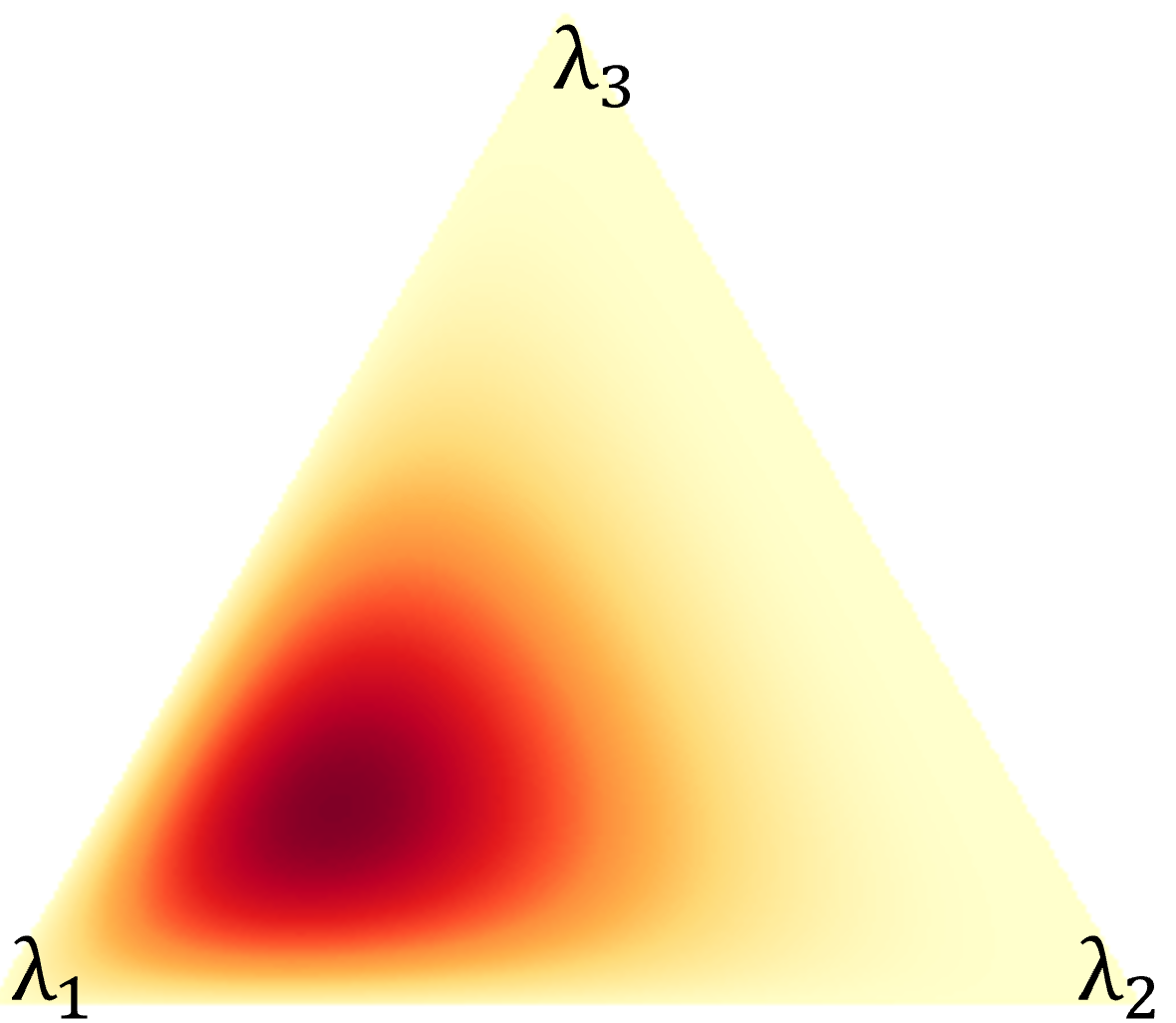}&
        \includegraphics[width=0.2\linewidth]{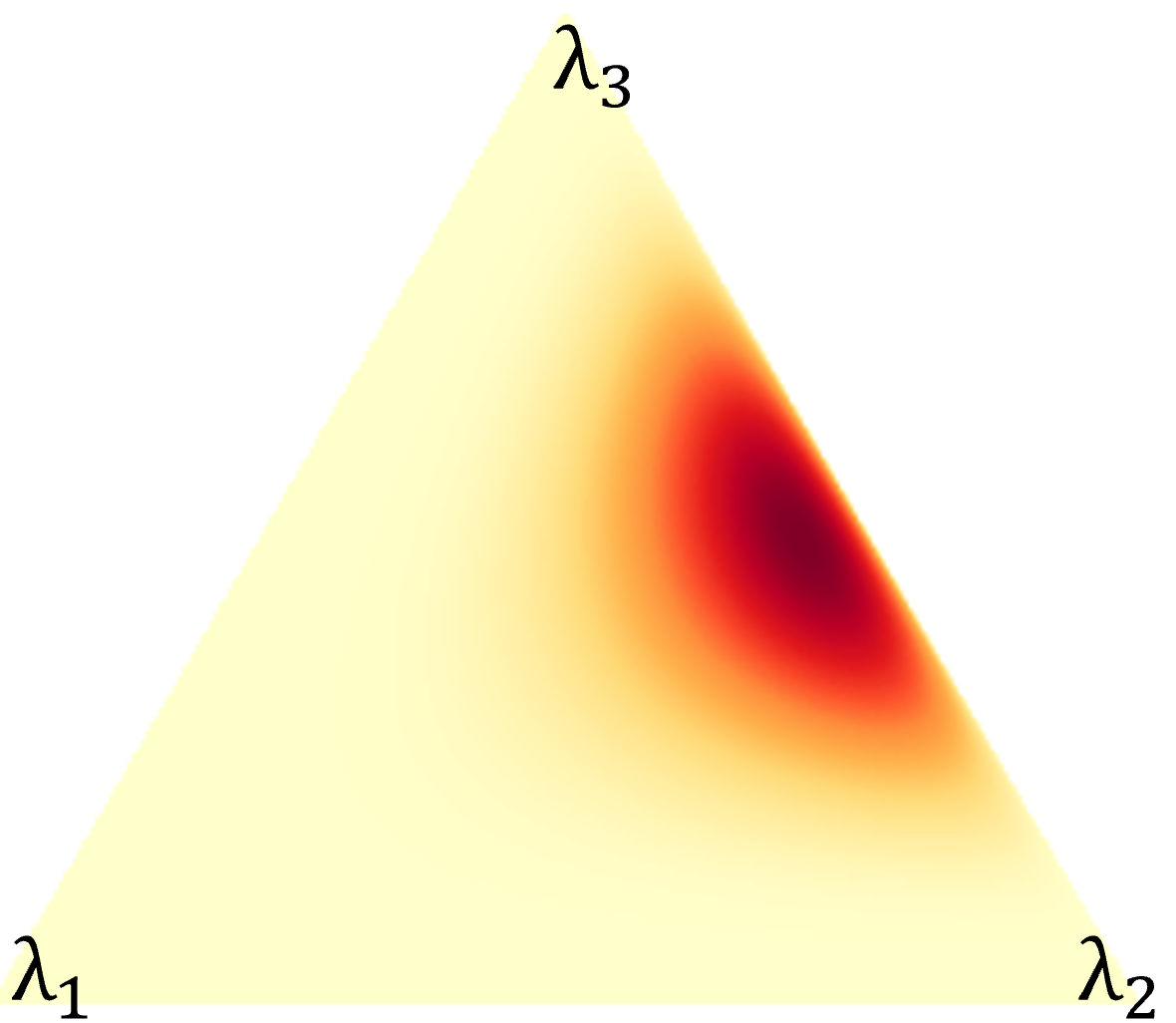}&
        \includegraphics[width=0.2\linewidth]{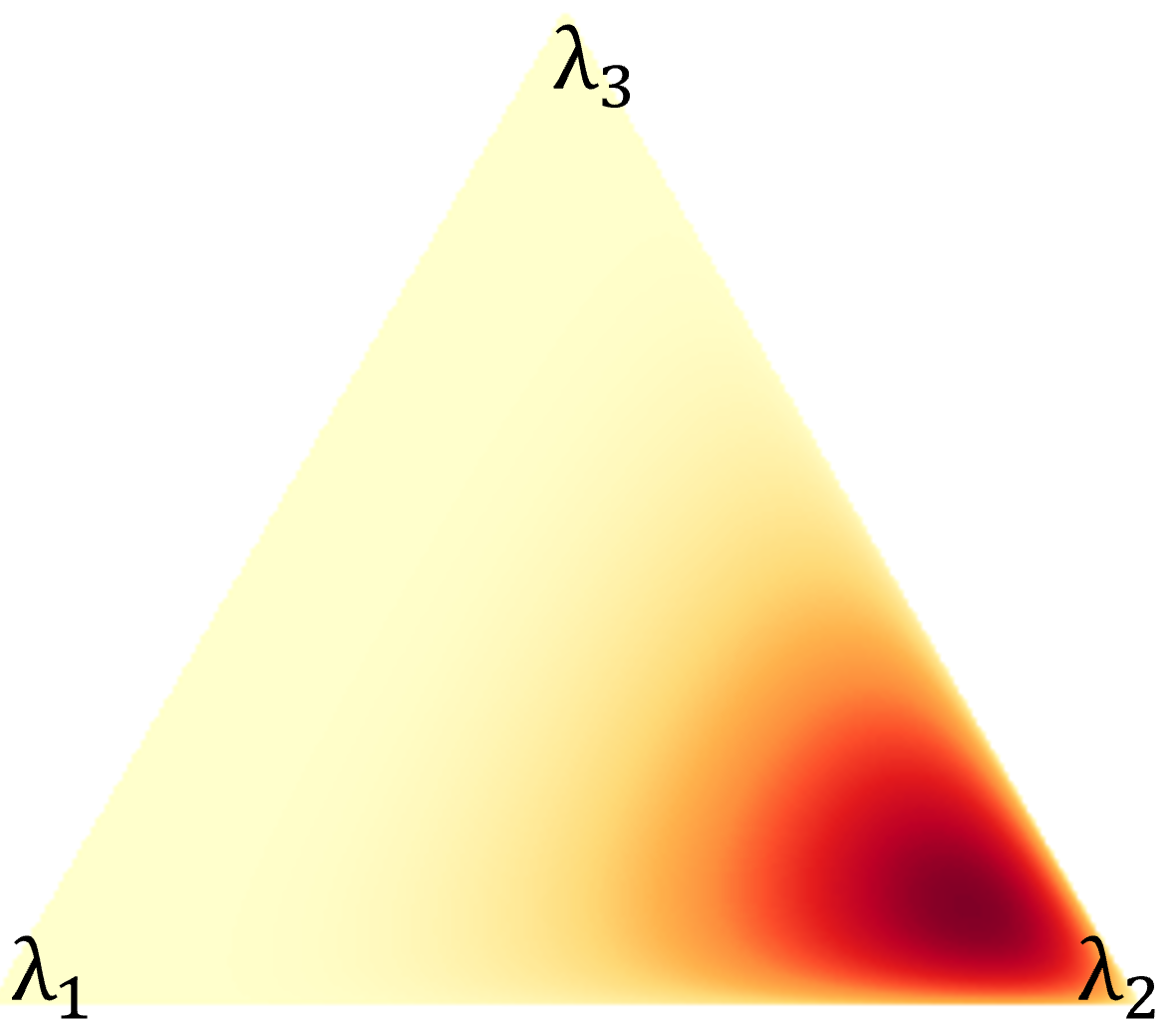}&
        \includegraphics[width=0.05\linewidth]{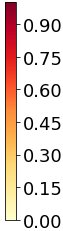}\\
        
        \rotatebox{90}{\hspace{0.52cm} \textbf{samples}} &
        \includegraphics[width=0.2\linewidth]{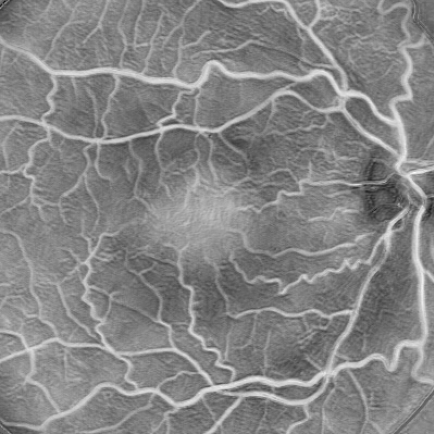}&  
        \includegraphics[width=0.2\linewidth]{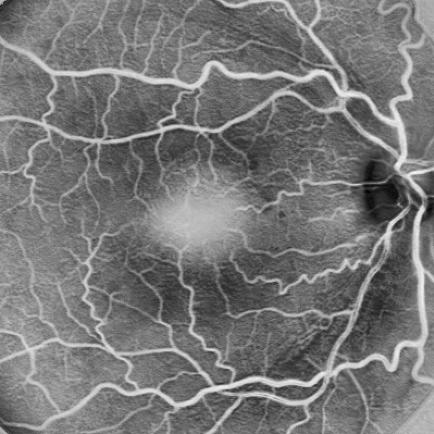}&
        \includegraphics[width=0.2\linewidth]{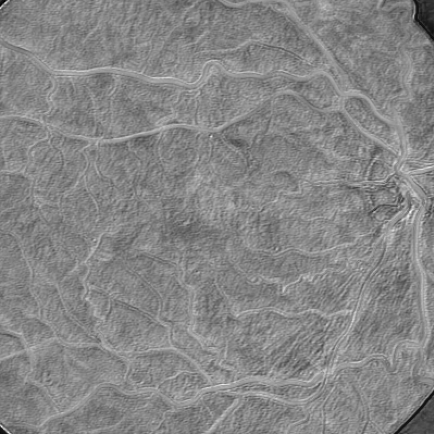}&
        \includegraphics[width=0.2\linewidth]{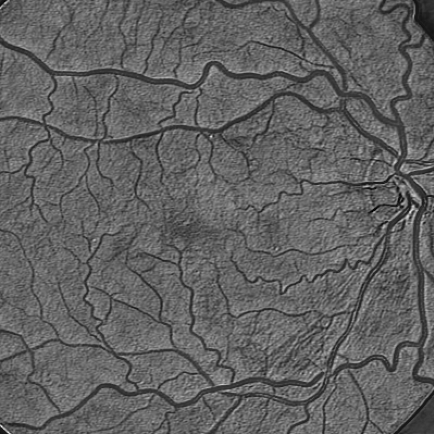}&\\
    \end{tabular}
    \caption{Examples of Dirichlet distribution and corresponding sample images.}
    \label{fig:dirichlet}
\end{figure}

In order to mixup three examples, we set a coefficient vector $\boldsymbol\lambda$ follow the Dirichlet distribution, i.e., $\boldsymbol\lambda\sim \text{Dirichlet}(\boldsymbol\alpha)$ where $\boldsymbol\lambda,\boldsymbol\alpha \in \mathbb{R}^3$. The probability density function (PDF) is defined as follows:

\begin{equation}
P(\boldsymbol\lambda)=\frac{\Gamma(\alpha_0)}{\Gamma(\alpha_1)\Gamma(\alpha_2)}\prod_{i=1}^3\lambda_{i}^{\alpha_i-1}\mathds{1}(\boldsymbol\lambda\in H),    
\end{equation}
with $H=\{\boldsymbol\lambda\in\mathbb{R}^3:\lambda_i\geq 0, \sum_{i=1}^3\lambda_i=1\}$ and $\Gamma(\alpha_i)=(\alpha_i-1)!$. Examples of PDFs with different hyperparameters $\boldsymbol\alpha$ are shown in the top row of Fig.\ \ref{fig:dirichlet}. 

The mixup image $\mathbf{s}_i$ is created by sampling the coefficient vector $\boldsymbol\lambda$ from $P(\boldsymbol\lambda)$, i.e., $\mathbf{s}_i=\lambda_1\mathbf{x}_i^0+\lambda_2\mathbf{x}_i^2+\lambda_2\mathbf{x}_i^3$. It is evident from the bottom row of Fig.\ \ref{fig:data_example} that the samples drawn from different distributions drastically vary in terms of contrast and vessel intensity. Thus, the Dirichlet mixup can augment the training data with varying styles of images without altering the vessel structure. To thoroughly exploit the continuous image space, we set $\boldsymbol\alpha=[1,1,1]$  such that $P(\boldsymbol\lambda)$ is a uniform distribution and all samples are considered equally. 

\subsection{Structural Correlation Constraints} \label{sec:losses}
\begin{figure}[t]
    \setlength{\tabcolsep}{10pt}
    \centering
    \begin{tabular}{cc}
        \includegraphics[width=0.22\linewidth]{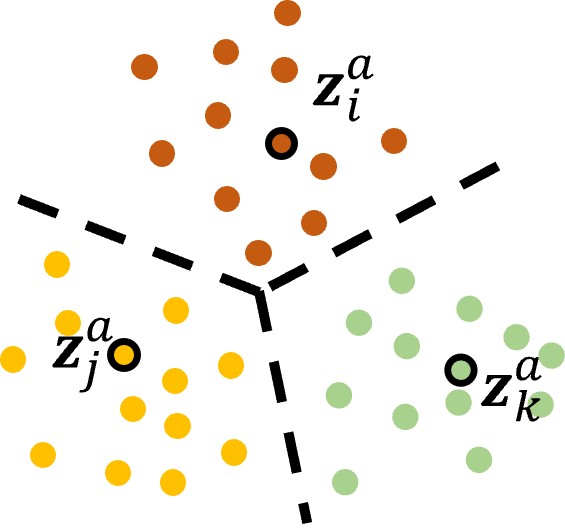} &
        \includegraphics[width=0.26\linewidth]{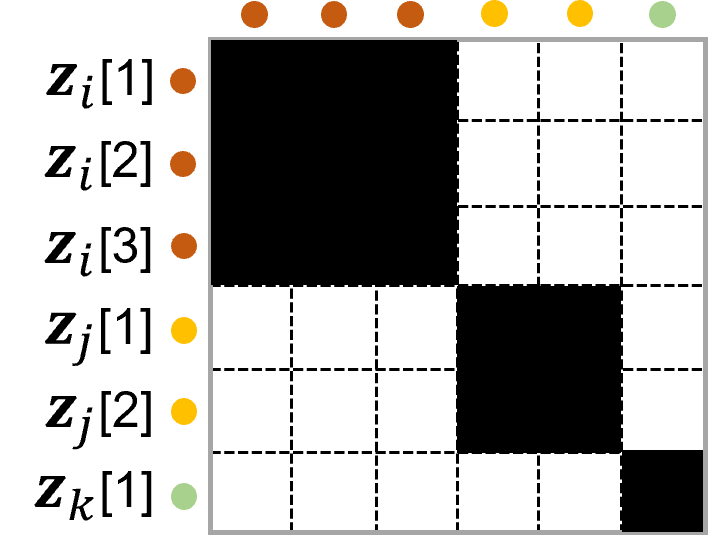}
    \end{tabular}
    \caption{\textbf{Left}: Feature clusters. Each dot represents a feature vector. Samples representing different anatomies are shown in different colors. The highlighted dots are the latent anchor features extracted from $\mathbf{x}_i^1$, $\mathbf{x}_j^1$ and $\mathbf{x}_k^1$ during meta-training. \textbf{Right}: NCC matrix. Each entry of the matrix is the cross-correlation between two feature vectors.}
    \label{fig:losses}
\end{figure}

Next, we design constraints to facilitate the model's concentration on the vessel morphology. We tackle this by delineating the correlation between latent features, as illustrated in Fig.\ \ref{fig:pipeline}(d). For two input images $\mathbf{x}_i$ and $\mathbf{x}_j$ ($i\neq j$), the features $\mathbf{z}_i$ and $\mathbf{z}_j$ are desired to be far apart, as their anatomies differ. In contrast, the $M$ mixup samples $\mathbf{s}_i^{(m)}$ for  $m\in\{1,\cdots,M\}$ are all anatomy-consistent, thus the corresponding features $\mathbf{z}_i^{(m)}$ should form subject-specific clusters, as shown in Fig.\ \ref{fig:losses}(left). Based on this intuition, we propose two loss functions.

\noindent
\underline{\normalfont{\textbf{Similarity loss $\mathcal{L}_{sim}$.}}} As mentioned in Sec.\ \ref{sec:metalearning}, we set $\mathcal{S}_{train}=\mathcal{D}^1$. The feature vector extracted during meta-training can be regarded as an anchor in the latent space; we denote it as $\mathbf{z}^a_i$. Then the latent features $\mathbf{z}_i^{(m)}$ from samples $\mathbf{s}_i^{(m)}$, $m\in\{1,\cdots,M\}$, should be close to the anchor $\mathbf{z}^a_i$. Here, we simply use the L1 norm as the similarity loss $\mathcal{L}_{sim}=\sum_{i=1}^N\sum_{m=1}^M\|\mathbf{z}_i^{(m)}-\mathbf{z}^a_i\|_1$,
where $N$ is the number of input images. $\mathcal{L}_{sim}$ is used to reduce the distance between sample features and the anchor within the clusters, as shown in Fig.\ \ref{fig:losses}(left).

\noindent
\underline{\normalfont{\textbf{Normalized cross-correlation loss $\mathcal{L}_{ncc}$.}}} In the context of contrastive learning, the Barlow Twins objective function \cite{zbontar2021barlow} was proposed to minimize the redundant information contained in the embedding vectors. This is realized by computing an empirical cross-correlation matrix of two vectors and bringing it closer to identity such that unmatched entries are not correlated. We extend this idea to a stack of vectors, as illustrated in Fig.\ \ref{fig:losses}(right). Feature vectors are color coded in the same way as the left panel of the figure. The normalized cross-correlations (NCC) between each pair of features form a symmetric matrix $\mathcal{C}$. As an example, the NCC of $\mathbf{z}_i^{(3)}$ and $\mathbf{z}_j^{(2)}$:
\begin{equation}
\mathcal{C}_{3,5}=\mathcal{C}_{5,3}=\frac{\mathbf{z}_i^{(3)}\cdot\mathbf{z}_j^{(2)}}{\sqrt{\mathbf{z}_i^{(3)}\cdot\mathbf{z}_i^{(3)}}\sqrt{\mathbf{z}_j^{(2)}\cdot\mathbf{z}_j^{(2)}}}
\end{equation}
In the ideal ground truth $\mathcal{C}^{*}$, the entries in the black region are 1, indicating similar features. Conversely, the white region entries are 0, representing dissimilarity. Then the NCC loss can be defined by $\mathcal{L}_{ncc} = \|\mathcal{C}^{*}-\mathcal{C}\|_{F}^2$.

The total loss for the meta-test stage is $\mathcal{L}_{test}=\omega_1\mathcal{L}_{seg}+\omega_2\mathcal{L}_{sim}+\omega_3\mathcal{L}_{ncc}$. Empirically, we set $\omega_1=\omega_2=100$, $\omega_3=1$.

\subsection{Experimental settings}

\noindent
\underline{\normalfont{\textbf{Datasets.}}} We use 7 public datasets listed in Table \ref{tab:dataset}.
The source domain $\mathcal{S}$ includes  three color fundus datasets: DRIVE, STARE and healthy samples in ARIA. By testing on the target domain $\mathcal{T}$, we evaluate the model's ability to generalize across pathological, cross-site, and cross-modality shift conditions.   

\noindent
\underline{\normalfont{\textbf{Implementation Details.}}} The segmentation network $g(\cdot)$ is a 6-layer residual U-Net. If the number of channels $n$ for a layer is denoted as $C_n$, then the architecture is: $C_{8}-C_{32}-C_{32}-C_{64}-C_{64}-C_{16}$. The synthesis model $f(\cdot)$ only functions on color fundus images in $\mathcal{S}$ during training. At test-time, fundus images are converted to grayscale by applying CLAHE on intensity-reversed green channel, while OCT-A and FA images are passed to the segmentation network $g(\cdot)$ directly. $g(\cdot)$ is trained and tested on an NVIDIA RTX 2080TI 11GB GPU. We set the batch size to 10 and train for 30 epochs. We utilize the Adam optimizer with the initial learning rate $\eta_{train} = 1\times 10^{-3}$ for meta-training and $\eta_{test} = 5\times 10^{-3}$ meta-testing, both decayed by 0.5 for every 3 epochs.

\input{table1.tex}
\input{table2.tex}

\section{Results}

\input{table3.tex}

\noindent
\underline{\normalfont{\textbf{Ablation Study.}}} In Table \ref{tab:ablation}, we investigate the contribution of the three major components of the proposed method: the episodic training paradigm, the similarity loss $\mathcal{L}_{sim}$ and the normalized cross-correlation loss $\mathcal{L}_{ncc}$. Note that $\mathcal{L}_{sim}$ requires the access to the latent anchor and thus is only applicable when using meta-training strategy. Without $\mathcal{L}_{sim}$ and $\mathcal{L}_{ncc}$, the model is trained with only the segmentation loss $\mathcal{L}_{seg}$. Our results show that the introduction of the episodic training provides noticeable improvement in all types of distribution shift. Both loss functions also contribute positively in general, and the proposed method ranks the best in types \RomanNumeralCaps{2} and \RomanNumeralCaps{3}, and second best in type \RomanNumeralCaps{1}.

\noindent
\underline{\normalfont{\textbf{Comparison to Competing Methods.}}} There are three major classes of approaches to solve the DG problem: data augmentation, domain alignment, and meta-learning. We compare against a representative algorithm from each: BigAug \cite{zhang2020generalizing}, domain regularization network \cite{aslani2020scanner}, and MASF \cite{dou2019domain}, respectively. We also compare to  VFT  \cite{hu2022domain} as it also focuses on leveraging shape information and pseudo-modalities. Moreover, we train a residual U-Net on $\mathcal{S}$ as a baseline model, and a residual U-Net on each target domain $T^p \in \mathcal{T}$ as an oracle model, to provide an indication of the lower and upper bounds of generalization performance. 

Table \ref{tab:result} compares the Dice coefficients (\%) of the competing methods. MAP ranks the best in almost all target domains (except RECOVERY, where it ranks second), which proves that the proposed MAP algorithm effectively enhances the robustness of the model under all three domain shift conditions. For some of the datasets such as ROSE and the diabetic subset of ARIA, the MAP's performance approaches the oracle. Compared to the VFT which explicitly models the tubular vessel shape, the implicit constraints provide a better guidance for the deep model to learn the structural features.

\section{Conclusion}
We present MAP, a method that approaches the DG problem by implicitly encouraging the model to learn about the vessel structure, which is considered to be a domain-agnostic feature. This is achieved by providing the model with synthesized images that have consistent vasculature but with significant variations in style. Then by setting constraints with regard to the correlation between latent features, the model is able to focus more on the target vessel structure. Our model's generalization capability is assessed on test data with different sources of domain shift, including data with pathological phenotypes, cross-site shifts, and cross-modality shifts.  The results indicate that the proposed method can greatly improve the robustness of the deep learning models across all three domain shift configurations.  

\noindent
\normalfont{\textbf{Acknowledgements.}} This work is supported by the NIH grant R01EY033969 and the Vanderbilt University Discovery Grant Program.

\clearpage
%
\bibliographystyle{splncs04}
\bibliography{refs.bib}

\end{document}

%% file: table1.tex
\begin{table}[t]
\centering
\begin{tabular}{p{0.25\textwidth}>{\centering}
                p{0.15\textwidth}>{\centering}
                p{0.20\textwidth}>{\centering\arraybackslash}
                p{0.12\textwidth}>{\centering\arraybackslash}
                p{0.12\textwidth}}
\specialrule{.1em}{.05em}{.05em}
\scriptsize{\hspace{2em}\textbf{dataset}} & \scriptsize{\textbf{modality}} & \scriptsize{\textbf{resolution}} & \scriptsize{\textbf{number}} & \scriptsize{\textbf{domain}} \\
\hline
\scriptsize{DRIVE \cite{staal2004ridge}} & \scriptsize{fundus} & \scriptsize{$565\times 584$} & \scriptsize{20} & \scriptsize{$\mathcal{S}$}\\
\scriptsize{STARE \cite{hoover2000locating}} & \scriptsize{fundus} & \scriptsize{$700\times 605$} & \scriptsize{20} & \scriptsize{$\mathcal{S}$}\\
\scriptsize{ARIA\cite{farnell2008enhancement} healthy} & \scriptsize{fundus} & \scriptsize{$768\times 576$} & \scriptsize{$61$} & \scriptsize{$\mathcal{S}$}\\
\rowcolor{Gray1}
\scriptsize{\hspace{3.9em}AMD} & \scriptsize{fundus} & \scriptsize{$768\times 576$} & \scriptsize{$59$} & \scriptsize{$\mathcal{T}$}\\
\rowcolor{Gray1}
\scriptsize{\hspace{3.9em}diabetic} &\scriptsize{fundus} & \scriptsize{$768\times 576$} &\scriptsize{$23$} & \scriptsize{$\mathcal{T}$}\\
\rowcolor{Gray2}
\scriptsize{{PRIME-FP20} \cite{ding2020weakly}} & \scriptsize{fundus} & \scriptsize{$4000\times 4000$} & \scriptsize{$15$} & \scriptsize{$\mathcal{T}$}\\
\rowcolor{Gray3}
\scriptsize{{ROSE} \cite{ma2020rose}} & \scriptsize{OCT-A} & \scriptsize{$304\times 304$} & \scriptsize{30} & \scriptsize{$\mathcal{T}$}\\
\rowcolor{Gray3}
\scriptsize{{OCTA-500(6M)} \cite{li2020image}} & \scriptsize{OCT-A} & \scriptsize{$400\times 400$} & \scriptsize{300} & \scriptsize{$\mathcal{T}$}\\
\rowcolor{Gray3}
\scriptsize{{RECOVERY-FA19} \cite{ding2020novel}} & \scriptsize{FA} & \scriptsize{$3900\times 3072$} & \scriptsize{8} & \scriptsize{$\mathcal{T}$}\\
\specialrule{.1em}{.05em}{.05em}
\end{tabular}
\caption{Datasets. Rows indicating the source domains have a white background while the target domains are shaded according to domain shift type. From top to bottom, (\RomanNumeralCaps{1}) pathology: light gray, (\RomanNumeralCaps{2}) cross-site: medium gray, (\RomanNumeralCaps{3}) cross-modality: dark gray.}
\label{tab:dataset}
\end{table}

%% file: table2.tex
\renewcommand{\arraystretch}{0.95}
\newcolumntype{a}{>{\columncolor{Gray1}}p}
\newcolumntype{b}{>{\columncolor{Gray2}}p}
\newcolumntype{d}{>{\columncolor{Gray3}}p}
\begin{table}[b]
    \centering
    \begin{tabular}{p{0.1\textwidth}>{\centering\arraybackslash}
                    p{0.1\textwidth}>{\centering\arraybackslash}
                    p{0.1\textwidth}>{\centering\arraybackslash}
                    a{0.14\textwidth}>{\centering\arraybackslash}
                    b{0.14\textwidth}>{\centering\arraybackslash}
                    d{0.14\textwidth}>{\centering\arraybackslash}
                    p{0.14\textwidth}}
        \specialrule{.1em}{.05em}{.05em}
        \scriptsize Episodic & $\scalemath{0.9}{\mathcal{L}_{sim}}$ & $\scalemath{0.9}{\mathcal{L}_{ncc}}$ & \scriptsize Type \RomanNumeralCaps{1} & \scriptsize Type \RomanNumeralCaps{2} & \scriptsize Type \RomanNumeralCaps{3} & \scriptsize Average\\
        
        \specialrule{.1em}{.05em}{.05em}

        \hspace{1.2em}\textbf{-} & \textbf{-} & \textbf{-} & $\scalemath{0.9}{62.93}$ & $\scalemath{0.9}{60.04}$ & $\scalemath{0.9}{63.94}$ & $\scalemath{0.9}{62.95}$\\

        \hspace{1.2em}\textbf{-} & \textbf{-} & \cmark & $\scalemath{0.9}{64.73}$ & $\scalemath{0.9}{62.48}$ & $\scalemath{0.9}{68.06}$ & $\scalemath{0.9}{66.02}$\\

        \hspace{1.1em}\cmark & \textbf{-} & \textbf{-} & $\scalemath{0.9}{\bf{67.50}}$ & $\scalemath{0.9}{63.40}$ & $\scalemath{0.9}{64.25}$ & $\scalemath{0.9}{65.19}$\\

        \hspace{1.1em}\cmark & \cmark & \textbf{-} & $\scalemath{0.9}{64.75}$ & $\scalemath{0.9}{66.24}$ & $\scalemath{0.9}{68.30}$ & $\scalemath{0.9}{66.77}$\\

        \hspace{1.1em}\cmark & \textbf{-} & \cmark & $\scalemath{0.9}{66.10}$ & $\scalemath{0.9}{\underline{66.99}}$ & $\scalemath{0.9}{\underline{69.71}}$ & $\scalemath{0.9}{\underline{68.05}}$\\

        \hspace{1.1em}\cmark & \cmark & \cmark & $\scalemath{0.9}{\underline{67.39}}$ & $\scalemath{0.9}{\bf{66.99}}$ & $\scalemath{0.9}{\bf{71.60}}$ & $\scalemath{0.9}{\bf{69.43}}$\\
        
        \specialrule{.1em}{.05em}{.05em}
    \end{tabular}
    \caption{The ablation study on the main components of MAP on data with three types of distribution shift. Boldface: best result, underline: second-best result.}
    \label{tab:ablation}
\end{table}

%% file: table3.tex
\renewcommand{\arraystretch}{0.95}
\newcolumntype{a}{>{\columncolor{Gray1}}p}
\newcolumntype{b}{>{\columncolor{Gray2}}p}
\newcolumntype{d}{>{\columncolor{Gray3}}p}
\begin{table}[t]
    \centering
    \begin{tabular}{p{0.12\textwidth}>{\centering}
                    a{0.12\textwidth}>{\centering\arraybackslash}
                    a{0.12\textwidth}>{\centering\arraybackslash}
                    b{0.15\textwidth}>{\centering\arraybackslash}
                    d{0.12\textwidth}>{\centering\arraybackslash}
                    d{0.12\textwidth}>{\centering\arraybackslash}
                    d{0.15\textwidth}}
        \specialrule{.1em}{.05em}{.05em}
        \hspace{0.1em} \multirow{2}{*}{\scriptsize Method} &  \multicolumn{2}{c}{\cellcolor{Gray1}{\scriptsize ARIA}} & & & & \\
        
         & \scriptsize amd & \scriptsize diabetic & \multirow{-2}{*}{\scriptsize PRIME-FP20} & \multirow{-2}{*}{\scriptsize OCTA 500} & \multirow{-2}{*}{\scriptsize ROSE} & \multirow{-2}{*}{\hspace{-0.2em}\scriptsize RECOVERY}\\
         
        \specialrule{.1em}{.05em}{.05em}
        \scriptsize \hspace{0.08em} \textit{baseline} & $\scalemath{0.9}{63.82}$ & $\scalemath{0.9}{65.19}$ & $\scalemath{0.9}{47.31}$ & $\scalemath{0.9}{73.16}$ & $\scalemath{0.9}{67.41}$ & $\scalemath{0.9}{51.25}$\\
        \hline
        \scriptsize \hspace{0.1em} Regular\cite{aslani2020scanner} & $\scalemath{0.9}{64.89}$ & $\scalemath{0.9}{66.97}$ & $\scalemath{0.9}{55.76}$ & $\scalemath{0.9}{73.54}$ & $\scalemath{0.9}{68.36}$ & $\scalemath{0.9}{55.20}$\\
        \scriptsize \hspace{0.01em} BigAug\cite{zhang2020generalizing} & $\scalemath{0.9}{\underline{65.55}}$ & $\scalemath{0.9}{67.27}$ & $\scalemath{0.9}{59.97}$ & $\scalemath{0.9}{76.88}$ & $\scalemath{0.9}{69.32}$ & $\scalemath{0.9}{\bf{63.20}}$\\
        \scriptsize \hspace{0.2em} MASF\cite{dou2019domain} & $\scalemath{0.9}{65.33}$ & $\scalemath{0.9}{\underline{67.75}}$ & $\underline{\scalemath{0.9}{65.96}}$ & $\scalemath{0.9}{77.65}$ & $\scalemath{0.9}{67.25}$ & $\scalemath{0.9}{50.74}$\\
        \scriptsize \hspace{0.4em} VFT\cite{hu2022domain} & $\scalemath{0.9}{61.81}$ & $\scalemath{0.9}{64.05}$ & $\scalemath{0.9}{54.64}$ & $\scalemath{0.9}{\underline{77.91}}$ & $\scalemath{0.9}{\underline{72.81}}$ & $\scalemath{0.9}{48.28}$\\
        \scriptsize\hspace{0.6em} MAP & $\scalemath{0.9}{\bf{66.69}}^{\sim}$ & $\scalemath{0.9}{\bf{68.08}}^{\sim}$ & $\scalemath{0.9}{\bf{68.21}}^{\dagger}$ & $\scalemath{0.9}{\bf{78.71}}^{\dagger}$ & $\scalemath{0.9}{\bf{74.25}}^{\dagger}$ & $\scalemath{0.9}{\underline{61.85}}^{\dagger}$\\
        \hline
        \scriptsize \hspace{0.5em} \textit{oracle} & $\scalemath{0.9}{73.34}$ & $\scalemath{0.9}{70.65}$ & $\scalemath{0.9}{77.80}$ & $\scalemath{0.9}{86.57}$ & $\scalemath{0.9}{76.03}$ & $\scalemath{0.9}{74.54}$\\
        \specialrule{.1em}{.05em}{.05em}
    \end{tabular}
    \caption{The Dice values (\%) for testing on target domains. Boldface: best result, underline: second best result. $^\sim: \text{p-value} \geq 0.05$, $^\dagger: \text{p-value} \ll 0.05$ in paired t-test compared to the baseline. The background is encoded the same way as Table \ref{tab:dataset}.}
    \label{tab:result}
\end{table}